\documentclass[sigconf, nonacm]{acmart}
\usepackage{algorithm}
\usepackage{algorithmic}

\graphicspath{{figures/}}

\begin{document}

\title{Distribution-Informed Adaptation for kNN Graph Construction}

\author{Min Shaojie}
\orcid{0000-0003-3663-1026}
\affiliation{%
  \institution{Chongqing University}
  \streetaddress{174 Shazheng Street}
  \city{Chongqing}
  \postcode{400044}
  \country{China}
}
\email{alexmin@cqu.edu.cn}

\author{Liu Ji}
\affiliation{%
  \institution{Chongqing University}
  \streetaddress{174 Shazheng Street}
  \city{Chongqing}
  \postcode{400044}
  \country{China}
}
\email{liujiboy@cqu.edu.cn}

%

\begin{abstract}
  Graph-based kNN algorithms have garnered widespread popularity for machine learning tasks due to their simplicity and effectiveness. However, as factual data often inherit complex distributions, the conventional kNN graph's reliance on a unified k-value can hinder its performance. A crucial factor behind this challenge is the presence of ambiguous samples along decision boundaries that are inevitably more prone to incorrect classifications. To address the situation, we propose the \textbf{D}istribution-Informed \textbf{a}daptive k\textbf{NN G}raph (DaNNG), which combines adaptive kNN with distribution-aware graph construction. By incorporating an approximation of the distribution with customized k-adaption criteria, DaNNG can significantly improve performance on ambiguous samples, and hence enhance overall accuracy and generalization capability. Through rigorous evaluations on diverse benchmark datasets, DaNNG outperforms state-of-the-art algorithms, showcasing its adaptability and efficacy across various real-world scenarios.
\end{abstract}

\ccsdesc[500]{Networks~Topology analysis and generation}
\ccsdesc[500]{Computing methodologies~Machine learning approaches}
\ccsdesc[300]{Computing methodologies~Supervised learning by classification}

\keywords{Graph-Based Algorithm, Machine Learning, Adaptive kNN Graph}

\maketitle

\section{Introduction}
In the realm of data science and machine learning, the k-Nearest Neighbors (kNN) algorithm has emerged as a fundamental yet powerful tool for a wide range of classification and regression tasks. Its simplicity, single-parametric nature, and ability to capture local patterns make it a popular choice in various domains, including image recognition \cite{knn4image}, natural language processing \cite{knn4nlp}, and recommendation systems \cite{knn4recommend}. Despite its popularity and ease of implementation, the traditional kNN approach comes with inherent limitations that can affect its performance and applicability in certain scenarios.

One of the primary limitations of traditional kNN lies in its dependence on feature-based representations, potentially disregarding deep-level relationships and dependencies among data points. To address this limitation, researchers have explored the use of kNN graph (kNNG) \cite{fuEFANNAExtremelyFast2016, boutetBeingPreparedSparse2016, dengFastKMeansBased2018}, where each data point is connected to its $k$ nearest neighbors in the feature space. As a graph-based approach, kNNG opened the avenue for capturing data relationships via graph-theory methodologies that can uncover the inherent data connections more effectively.

However, in kNNG, the inherited unified-k across all data points can still result in compromised generalization capabilities, especially in scenarios where the underlying data distribution exhibits complex patterns. Researchers have shown that adaptively determining $k$ values can improve the algorithm's robustness. Adaptive methods for kNNG, as well as kNN, often adjust k-value under the guidance of certain criteria, such as regional accuracy \cite{lvknn} or local density \cite{aknng}. Local distribution information has also been introduced into the determination process, such as the adaKNN \cite{adaknn}, SMKNN \cite{smknn}, and DC-LAKNN \cite{dclaknn}, aiming to enhance the performance with the aid of the varying local density.

A critical challenge for cutting-edge adaptive kNNG algorithms stems from ambiguous samples. The ambiguous samples can be defined as:
\begin{definition}[Ambiguous Samples]
Ambiguous samples\footnote{Ambiguous samples will be refer to as ``borderline samples'' to highlight their characteristic.}, or borderline samples, refer to samples that lie close to the decision boundaries among different classes.
\end{definition}

Ambiguous samples' approximation to other classes makes them difficult to be predicted \cite{saezSMOTEIPFAddressing2015, nguyenBorderline2011}. Improving performance on borderline samples will naturally lead to a more satisfactory overall performance.

In this paper, we demonstrate that integrating an approximation of the original data distribution as a cohesive entity is naturally suitable to adaptively determine $k$ and solve the predicament posed by borderline samples. we present a novel approach called the \textbf{D}istribution-Informed \textbf{a}daptive k\textbf{NN G}raph (DaNNG) which incorporates distribution's approximation into the construction of a heterogeneous graph representation. The distribution-informed DaNNG learns an approximation of overall distribution, within which the $k$ values for borderline samples are adapted under guidance of customized criteria. It then leverages the acquired distribution-based information and traditional k-value to reach an optimal k-sequence for data. Under rigorous evaluations, we achieve significant performance gain on borderline samples, resulting in improved overall accuracy and generalization capability.

\section{Related Work}

The concept of adaptive-k in kNN algorithms focuses on dynamically determining the optimal value of $k$ for each data point, while graph-based kNN methods leverage graph structures to represent and analyze data relationships. Both adaptive kNN and graph-based kNN methods have shown promising results in improving the performance of traditional kNN.

\subsection{Adaptive kNN}
Adaptive kNN algorithms offer the advantage of dynamically determining the optimal k-value for each data point, allowing the model to adapt to varying data densities and achieve improved performance in complex and diverse datasets. One notable development in the area of adaptive kNN is the work of Ada-kNN2 by Mullick et al. \cite{adaknn2}. The author addressed the challenge of varying densities across data regions by learning suitable k-value from the test-point neighborhood with the help of artificial neural networks. Similarly, Jodas et al. presented the parameterless nearest neighbors classifier (PL-kNN) \cite{plknn}, where $k$ is automatically adjusted based on the underlying data characteristics, alleviating the need for manual parameter tuning. A tree-based adaptive kNN, named KTree, was developed by Zhang et al. \cite{ktree}, in which optimal-k can be efficiently obtained for test samples from a pre-constructed decision tree. The proposed KTree exhibits improved resilience to class imbalance, enhancing the robustness of kNN classifiers in challenging data scenarios. 

\subsection{Graph-Based kNN}
Graph-based kNN approaches (i.e., kNNG) leverage graph structures to represent data relationships, enabling a more intuitive and interpretable representation of data. One example is the concept of the centered kNNG introduced by Suzuki and Hara \cite{centered} for semi-supervised learning. Centered kNNG aims to reduce the presence of hubs by eliminating common directions of inner-product on the Hilbert space, which not only enhances robustness but also incurs minimal computational overhead.

Like adaptive kNN, kNN Graph can also adopt adaptive strategies to determine $k$ values, resulting in graphs exhibiting heterogeneity in node degrees. The AKNNG \cite{aknng} dynamically adjusts the number of neighbors with an enhanced similarity matrix to explore optimal-k and its application in spectral clustering. Murrugarra-Llerena and Andrade Lopes \cite{murrugarrallerenaAdaptiveGraphbasedKnearest2011} presented an adaptive graph-based kNN method that iteratively adjusts neighbor numbers for misclassified samples. Furthermore, Yan et al. \cite{yanRobustGravitationBased2022} introduced a robust gravitation-based approach addressing the challenges of class-imbalanced scenarios in kNNG.

Distribution's significance in the determination of $k$ for kNNG has been verified in previous works, including \cite{yanRobustGravitationBased2022, ktree, murrugarrallerenaAdaptiveGraphbasedKnearest2011}. These methods have demonstrated the advantages of considering data distribution as model inputs, leading to more accurate and robust representations of data relationships. However, existing approaches neglect the interrelation between borderline samples and data distribution, potentially limitting their ability to capture underlying relationships within the data. In contrast, our DaNNG integrates the data distribution while simultaneous considering the behaviour of borderline samples' $k$, substantially improving borderline-sample accuracy, and hence strengthening overall model robustness.

\section{Preliminaries} \label{sec:prerequisites}

\subsection{The k-Nearest Neighbors Graph (kNNG)}
The kNNG can be considered as a representation of kNN but with more flexible graph-based prediction strategies. For $n \times d$ data $X$ and a given $n \times n$ similarity matrix $S$, we generate a $n \times n$ binary matrix $W$ such that $W_{ij} = 1 \text{ iff. } j \in NN_k(X_i)$. A plain kNNG is an undirected graph $G=(V,E)$ in which $V_i$ represents the $i_{th}$ sample in $X$ and the corresponding adjacent matrix $A$ is determined as $A_{ij} = A_{ji} = 1 \text{ iff. } (W_{ij} = 1) \vee (W_{ji} = 1)$. For more complicated situations, a mutual kNNG indicates $A_{ij}^{mutual} = A_{ji}^{mutual} =1 \text{ iff. } (W_{ij} = 1) \wedge  (W_{ji} = 1)$, whereas a directed kNNG has $A_{ij}^{directed} = 1 \text{ iff. } W_{ij} = 1$.

\subsection{The Kullback-Leibler Divergence}
The Kullback-Leibler (KL) Divergence, also known as relative entropy, is a fundamental concept in information theory and statistics. Introduced by Solomon Kullback and Richard Leibler in the 1950s, KL divergence measures the dissimilarity between two probability distributions. The original KL divergence formula is given by:
\begin{equation} \label{eq:kld}
    D_{KL}(P || Q) = \sum_{i} p_i \log \frac{p_i}{q_i}
\end{equation}
where $P$ and $Q$ are two probability distributions.

In our model, The concept of KL divergence is exploited to measure the sample-wise estimation rather than the value-oriented distribution, aiming to acquire the fitness kernel $F$.

\subsection{The Fitness Kernel}
The fitness kernel, originally referred to as a mathematical function that quantifies the fitness or suitability of nodes in a network under the context of network science, is utilized to determine $k$ for each sample in our model.

Given a set of d-dimensional vectors $X = \{x_1, x_2, \cdots , x_n \} \in \mathbb{R}^{n\times d}$, we define the fitness kernel $F = \{f_1, f_2, \cdots , f_n \} \in \mathbb{R}^{n\times 1}$ as a one-dimensional vector, serving as an approximation of the data distribution with suitable variation.

\subsection{Research Objective}
Given a d-dimensional vector $X \in \mathbb{R}^{n\times d}$, our objective is to construct an adaptive kNNG, $G=(V,E)$, called the \textbf{D}istribution-Informed \textbf{a}daptive k\textbf{NN G}raph (DaNNG), aiming to effectively improve performance on borderline samples and thus enhance overall robustness.

\subsection{Quantified Borderline Samples}
To facilitate quantitative evaluation, borderline samples are quantified as follows:
\begin{definition}[i-Quantile Borderline]
    The i-Quantile Borderline, denoted as $iQB$, refers to a subset of samples whose probability estimations lie in the lower $i$ percent of the population.
\end{definition}

\subsection{Distribution-Informed Adaption} \label{sec:matters}

\begin{figure*}[ht]
  \centering
  \includegraphics[width=0.9\textwidth]{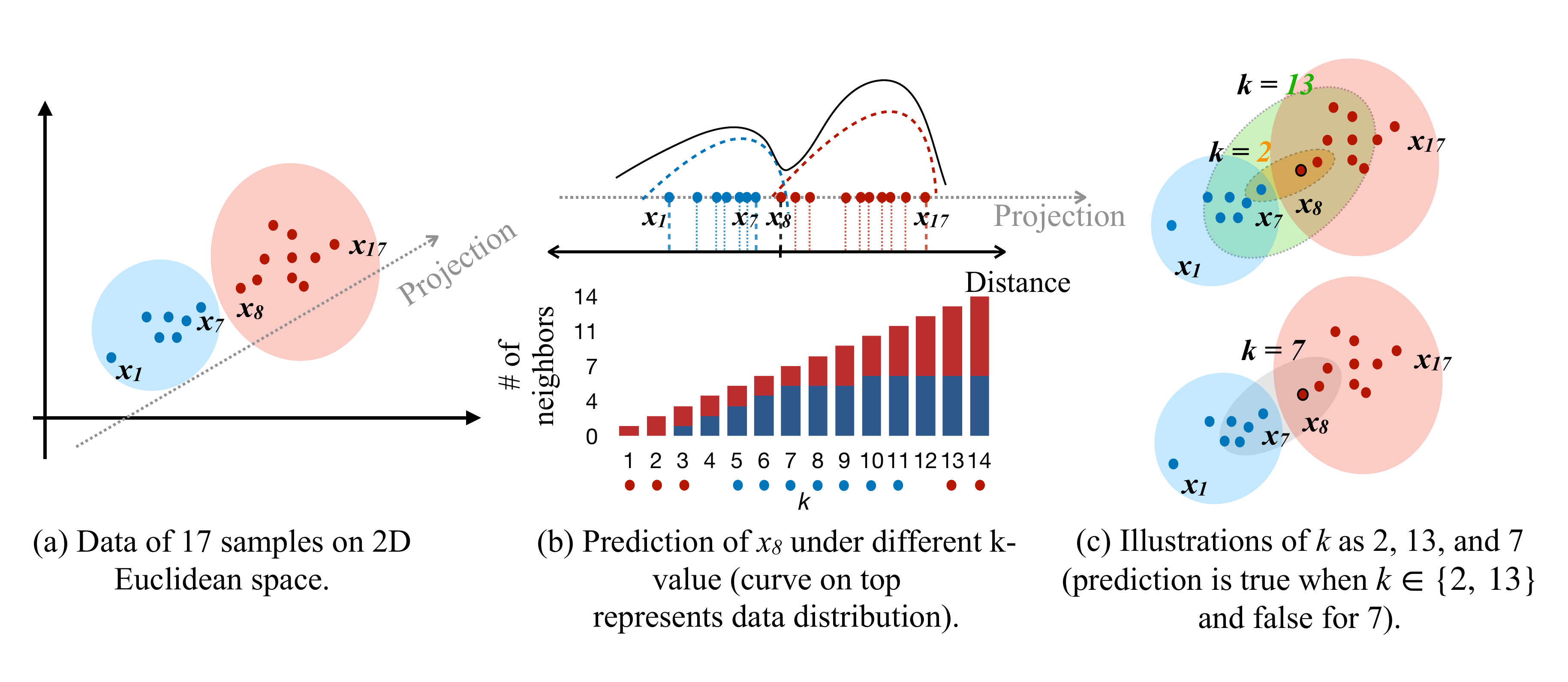}
  \caption{Demonstration of borderline-sample misclassification under data containing two left-skewed classes.}
  \Description{The prediction of borderline samples will be incorrect under normal k values due to the proximity of other classes, but can potentially become correct with relatively larger or smaller k.}
  \label{fig:overall_illustration}
\end{figure*}

In this section, we demonstrate why introducing overall distribution to the adaption of $k$ can be rewarding.

Assuming the subsamples within a classification task are several Gaussian-distributed clusters, a unified $k$ for samples on the cluster border will inevitably include samples from other classes, bringing uncertainty into prediction. To elaborate, we provide an intuitive illustration via two-class (blue and red) data $X = \{x_1, x_2, \cdots, x_{17} \}$ in Figure \ref{fig:overall_illustration}.

Under the projection direction, $x_1, x_7$ and $x_8, x_{17}$ are at the border of each class (i.e., borderline samples). If we try different values of $k$ to classify the borderline sample $x_8$, we will observe a false prediction (blue) under common $k$ values (from 5 to 11). This pattern is strengthened if different classes exhibit the same side of skewness (two classes in Figure \ref{fig:overall_illustration}(b) are both left-skewed). 

A viable solution is to assign larger/smaller $k$ for borderline samples, aiming to achieve the following:
\begin{itemize}
  \item {\texttt{Smaller-k}}: This choice favors a narrower range of samples, prioritizing the closest neighbors that are more likely to belong to the same class.
  \item {\texttt{Larger-k}}: Including a larger range of samples in the process helps to capture more comprehensive information.
\end{itemize}

In the provided example, it becomes evident that both opting for smaller-k and choosing larger-k can lead to accurate results. (predicted 'red' when $k \in \{1, 2, 3\} \text{ or } k \in \{13, 14\}$).

This pattern of the k-value, evolving $k$ to either larger or smaller for borderline samples, is reconciled with the distribution of data because no matter whether $k$ becomes larger or smaller the corresponding probability estimation will both become lower. In other words, patterns of low probability estimation for borderline samples align with the low probability estimations for their $k$ values after adaption, making the overall distribution information an ideal choice of input for k-determination.

\section{Methodology} \label{sec:formation}
The \textbf{D}istribution-Informed \textbf{a}daptive k\textbf{NN G}raph (DaNNG) model is presented in this section. The construction of DaNNG can be summarized as the adaptive determination of the value of $k$ for each data point with the pre-learned fitness kernel, $F$, which approximates the original distribution under the guidence of local topology on borderline samples.

\subsection{Determining k}

The $k$ values for samples in DaNNG are determined using the following formula:
\begin{equation} \label{eq:k}
    K = (1 - \eta)\kappa + \eta F + \epsilon    
\end{equation}
where $K \in \mathbb{R}^{n\times 1} $ represents the corresponding $k$ value for each sample, $\eta \in [0, 1]$ is the parameter controlling the trade-off between traditional $k$ (i.e., $\kappa$) and the fitness kernel $F$. Additionally, $\epsilon \in \mathbb{R}^{n\times 1}$ is a random vector in which $\epsilon_i \sim \mathcal{N}(0, {\frac{1}{3}}^2)$ accounting for stochastic variations that can only increase/decrease $k_i$ by 1 at most ($3\sigma = 1$).

\subsection{Acquiring the Fitness Kernel}

We aim to acquire a fitness kernel, F, that approximates the original distribution. In DaNNG, $F$ is obtained through a two-step process wherein the objective is to iteratively minimize sample-wise Kullback-Leibler divergence under customized constraints.

\begin{algorithm}[t]
    \caption{Acquire Fitness Kernel}
    \label{alg:fitness}
    \textbf{Input}: $X,\ y$ - Data matrix of size $(n, d)$ and prediction target of size $(n, 1)$\\
    \textbf{Output}: $F$ - Vector of fitness kernel in $(n, 1)$
    
    \begin{algorithmic}[1]
    \STATE Initialize $F$ with Equation \ref{eq:f_init}: $F_{\text{init}} = \log(\text{bincount}(y))$
    \STATE Let $F = F_{\text{init}}$ and $loss = 1$
    \STATE Compute sample-wise $X_{\text{probs}}$
    \STATE Set $threshold = 1e-2$
    \WHILE{$loss > threshold$}
        \STATE Compute sample-wise $F_{\text{probs}}$
        \STATE Compute $loss$ with Equation \ref{eq:loss}: \\ $loss = \text{KLDivergence}(X_{\text{probs}}, F_{\text{probs}}) \text{ s.t. } g(F) \le 0$
        \IF{$loss < threshold$}
            \STATE \textbf{break}
        \ENDIF
        \STATE Update $F$: $F -= learning\_rate \times loss$
    \ENDWHILE
    \STATE Normalize $F$: $F_{\text{scaled}} = \text{Scale}(F, \min=\frac{1}{2}\kappa, \max=\frac{3}{2}\kappa)$
    \RETURN $F_{\text{scaled}}$
    \end{algorithmic}
\end{algorithm}

\subsubsection{Initialization of $F$}

We initialize the fitness kernel $F$ by assigning equal values to each data point in the same classes, in which higher values are assigned to dense regions to achieve high level of clustering. This serves as the starting point for the iterative process of minimizing the KL divergence.

Given a dataset with $m$ classes of samples:
\begin{multline} \label{eq:inputs}
    X = \{x_1, x_2, \cdots , x_n \} \in \mathbb{R}^{n\times d}, y = \{y_1, y_2, \cdots ,\\
     y_n \in C \} \in \mathbb{R}^{n\times 1}
\end{multline}
where $C = \{c_1, c_2, \cdots, c_m\}$ represents target labels.

We initialize $F$ as:
\begin{equation} \label{eq:f_init}
    F_{\text{init}} = \log  N_y
\end{equation}
where $N_{y_i}$ is the sample count of class $y_i$, and the logarithm aims to prevent a cascade effect in which classes with large amounts of samples may result in a fully connected graph.

\subsubsection{Adaption Criteria for Borderline Samples} \label{sec:criteria}

\begin{figure}[t]
  \centering
  \includegraphics[width=0.9\columnwidth]{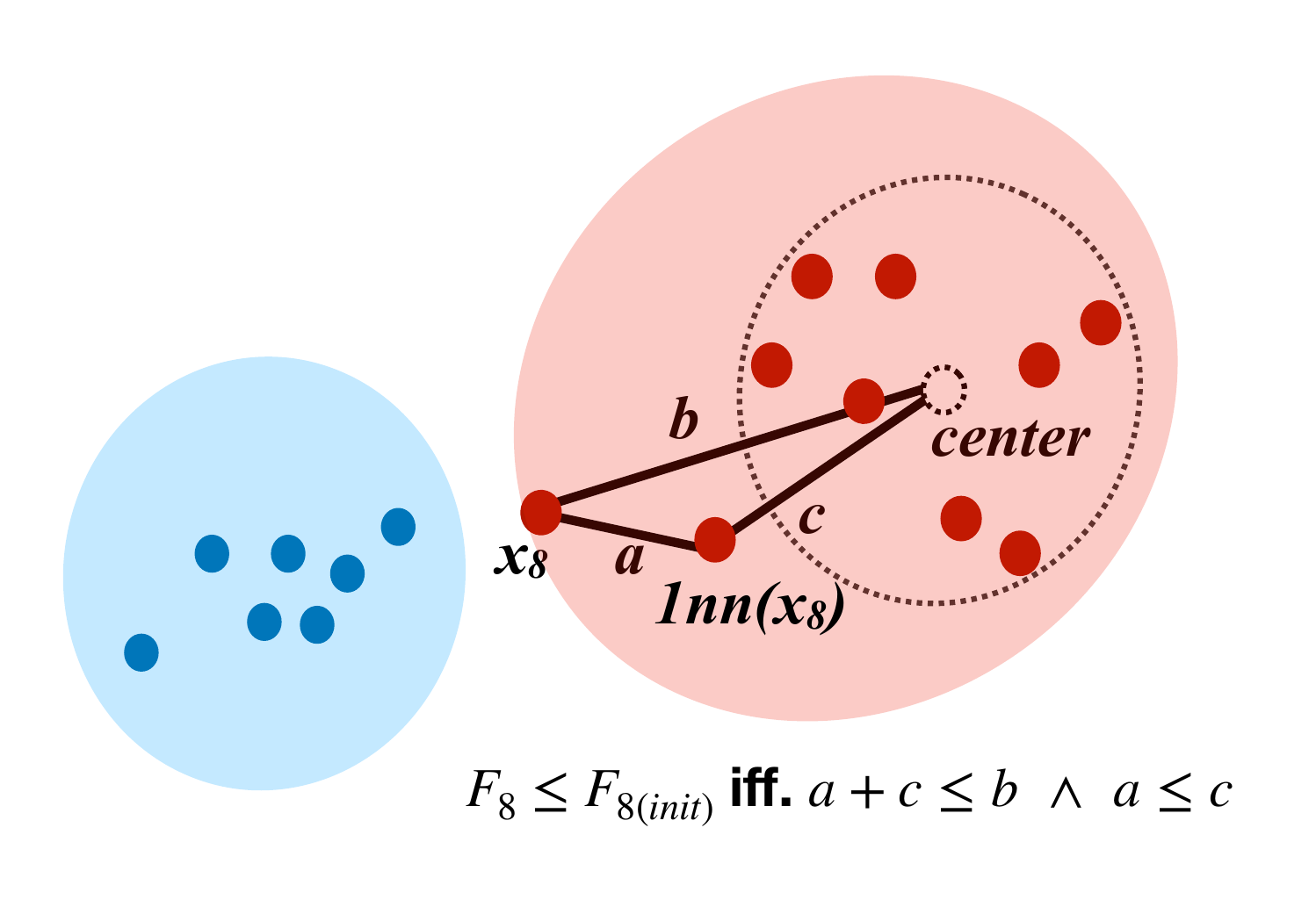}
  \caption{Criteria demonstration of circumstance when $F$ evolves to smaller value for borderline samples.}
  \Description{The value of F should become smaller during optimization for borderline samples if and only if 1. The shortest path to the center of their class (excluding the sample itself and its nearest neighbor) traverses through the nearest neighbor; and 2. The distance to their nearest neighbor is either equal to or less than the distance between the nearest neighbor and the class center.}
  \label{fig:path}
\end{figure}

As elucidated in Section \ref{sec:matters}, borderline samples exhibit two directions of evolution, one towards larger values of $k$ and the other towards smaller $k$, both of which can potential enhance model robustness. To discern which direction is the most appropriate for each sample in the process of optimizing KL Divergence, we devised a straightforward rule based on the strong correlation between the proximity of the nearest neighbor ($1\text{NN}$) and smaller $k$ (as elaborated in Section \ref{sec:matters}).

This rule is derived from the local topology and prescribes the adoption of a smaller $k$ exclusively for borderline sample that meet the following conditions:

\begin{itemize}
  \item The shortest path to the center of its class (excluding the sample itself and its nearest neighbor) traverses through the nearest neighbor (i.e., $a + c \le b$ in Figure \ref{fig:path}).
  \item The distance to their nearest neighbor is either equal to or less than the distance between the nearest neighbor and the class center (i.e., $a \le c$ in Figure \ref{fig:path}).
\end{itemize}

In Figure \ref{fig:path}, we have demonstrated the determination criteria of the smaller-$k$ evolution for $x_8$. This criteria for borderline samples can be formally represented as constraints during distribution approximation in the form:

\begin{equation} \label{eq:constraint}
\begin{array}{l}
  g(F) \le 0, 
  \text { s.t. } g_{i}(F)=\left\{\begin{array}{ll}
  f_{i}-f_{i} ^\text{init},  & i \in S \\
  f_{i} ^\text{init}-f_{i}, & i \in H 
  \end{array}\right.
\end{array}
\end{equation}
in which $S$ and $H$ are two subsets of the borderline set (i.e., $S,\ H \subseteq iQB$) that meet:

\begin{equation}
  \left\{\begin{array}{ll}
    \begin{aligned}
      x_i & \in S \text{ iff. } \text{dist}(x_i, 1\text{NN}(x_1)) < \text{dist}(1\text{NN}(x_1), \mu_i) \\
      & \wedge \text{dist}(x_i, 1\text{NN}(x_1)) + \text{dist}(1\text{NN}(x_1), \mu_i) \le \text{dist}(x_i, \mu_i),
    \end{aligned} \\
  H =  iQB - S
\end{array}\right.
\end{equation}
where $\mu_i$ is the class center of $y_i$ excluding $x_i$ and $1NN(x_i)$.

To assess the effectiveness of our proposed criteria, we have added a version of DaNNG during evaluations in Section \ref{sec:evaluations}.

\subsubsection{Approximate Original Distribution}

Since the KL divergence is originally intended for measuring similarities between two populations and is not suitable for comparing d-dimensional $X$ and one-dimensional $F$, we design a loss by modifying KL divergence to measure the dissimilarity of sample-wise probability between $X$ and $F$ instead:
\begin{equation} \label{eq:loss}
loss = \sum_{i=1}^{n} P(x_i) \cdot \log\left(\frac{{P(x_i)}}{{P(f_i)}}\right) \text{, s.t. } g(F) \le 0 
\end{equation}
where $P(x_i)$ and $P(f_i)$ are the probabilities for the $i$-th sample in $X$ and $F$, respectively.

The loss function is used to modify $F$ so that the fitness kernel preserves the original data distribution while the fitness values for borderline samples evolve to the intended directions under the guidance of constraint, $g(F) \le 0$. In DaNNG, we calculate the probability via Kernel Density Estimation (KDE), yielding the slope of gradient descent becomes:
\begin{multline} \label{eq:gradient}
    \mathrm{\nabla} = \frac{\partial loss}{\partial F} = \frac{1}{h^{2}\rho} \left( \rho \sum_{i=1}^{n} P(x_{i})(F-f_{i}) \right. \\
    + \left. P(X) \sum_{i=1}^{n} \rho_{i}(F-f_{i}) \right)
\end{multline}
where $\rho$ represents the KDE under Gaussian kernel, i.e., $\rho_i = {\textstyle \sum_{j=1}^{n}}e^{-\frac{1}{2 h^{2}}(f_i-f_j)^{2}}$ and $h$ is the bandwidth controlling distribution smoothing. Additionally, to ensure a consistent impact across different $\kappa$, $F$ is scaled proportionally to range $(\frac{1}{2}\kappa, \frac{3}{2}\kappa)$ (Proof and analysis provided in Section \ref{sec:analysis}.

\begin{algorithm}[t]
  \caption{DaNNG Construction}
  \label{alg:DaNNG}
  \textbf{Input}: $X,\ y, \text{ and } k$ - Data matrix of size $(n, d)$, prediction target of size $(n, 1)$ and user-defined $k$ of traditional kNN Graph\\
  \textbf{Output}: $A$ - Adjacency matrix $(n, n)$ of DaNNG
  
  \begin{algorithmic}[1]
  \STATE Compute  $F$ using Algorithm \ref{alg:fitness}
  
  \STATE Compute $\eta_m = \mathop{\arg\max}\limits_{\eta}(\text{Accuracy})$ \textbf{for} $\eta \in (0, 1)$
  \STATE Let $K = (1-\eta_m) \cdot \kappa + \eta_m \cdot F + \epsilon \sim \mathcal{N}(0, \sigma^2)$

  \STATE Set $W_\text{DaNNG} = \text{zeros}(n, n)$
  \FOR{each sample $i$}
      \STATE Compute $kNN_i = k_iNearestNeighbors(i)$
      \FOR{$j$ in $kNN_i$}
          \STATE Set the corresponding entries in $W_\text{DaNNG}$ to 1: $W^\text{DaNNG}_{(i, j)} = 1$
      \ENDFOR
  \ENDFOR
  \STATE $A_\text{DaNNG} = \max(W_\text{DaNNG}, W_\text{DaNNG}^T)$
  \RETURN $A_\text{DaNNG}$
  \end{algorithmic}
\end{algorithm}

\subsection{Graph Construction}
With $K$ determined and a given similarity measurement, the binary matrix $W$ for DaNNG is obtained as:
\begin{equation} \label{eq:matrix_w}
    W^\text{DaNNG}_{ij} = \begin{cases}
      1, & \text{if } j \in kNN_{k_i}(x_i) \\
      0, & \text{if } j \notin kNN_{k_i}(x_i)
    \end{cases}
\end{equation}

Using the same construction procedures described in Section \ref{sec:prerequisites}, the adjacent matrix $A$ for DaNNG is obtained as $A_{\text{DaNNG}} = \max(W_{\text{DaNNG}}, W_{\text{DaNNG}}^\mathrm{T})$, $A_{\text{DaNNG}}^{directed} = W_{\text{DaNNG}}$ for direct DaNNG, and $A_{\text{DaNNG}}^{mutual} = \min(W_{\text{DaNNG}}, W_{\text{DaNNG}}^\mathrm{T})$ for mutual DaNNG.

After the graph is constructed, various graph-based algorithms, such as label propagation \cite{propagation}, can be employed to make predictions.

\section{Theoretical Analysis} \label{sec:analysis}
In this section, we first provide how we solve the constrainted optimization in practice, and then establish that the scaling of $F$ is dispensable in DaNNG as long as it upholds symmetry in relation to $\kappa$.

\subsection{Constrained Optimization for KL Divergence}
With the pre-defined sample-wise KL divergence loss in Formula \ref{eq:loss}, we illustrate how we obtain the slope of the gradient loss in Equation \ref{eq:gradient}.

Since $P(X)$ is fixed, the optimization objective can be reformulated as:
\begin{equation} \label{eq:loss_re}
  \min_{F \in \mathbb{R}^{n\times 1} } \sum_{i=1}^{n} P(x_i) \cdot \log\left(\frac{{P(x_i)}}{{P(f_i)}}\right)
  \text{, s.t. } g(F) \le 0
\end{equation}

\subsubsection{Gradient of Loss} \label{sec:gradient_proof}

In practice, with Gaussian kernel KDE as the probability measurement, we have:
\begin{equation} \label{eq:deduct_gradient1}
    \begin{split}
        \frac{\partial loss}{\partial F} &= -{\textstyle \sum_{i=1}^{n}}P\left(x_{i}\right)\log{{\textstyle \sum_{j=1}^{n}} K\left( f_{i}-f_{j}, h\right )} \\
        & =-{\textstyle \sum_{i=1}^{n}}P\left(x_{i}\right)\log{{\textstyle \sum_{j=1}^{n}}  e^{-\frac{1}{2 h^{2}}\left(f_{i}-f_{j}\right)^{2}}}
    \end{split}
\end{equation}
where $K(\cdot)$ and $h$ are the kernel and bandwidth of KDE, respectively.

For each projection of $F$, we have:

\begin{multline} \label{eq:deduct_gradient_combined}
    \begin{aligned}
        \frac{\partial loss}{\partial f_{m}} &= \frac{1}{h^{2}} ( \sum_{i=1}^{n} P(x_{i})(f_{m}-f_{i}) \\
        & \ \ \ + P(x_{m}) \frac{ \sum_{i=1}^{n} (f_{m}-f_{i}) e^{-\frac{1}{2 h^{2}}(f_{m}-f_{i})^{2}}}{ \sum_{i=1}^{n}  e^{-\frac{1}{2 h^{2}}(f_{m}-f_{i})^{2}}} ) \\
        &= \frac{1}{h^{2}\rho_m} ( \rho_m \sum_{i=1}^{n}  P(x_{i})(f_{m}-f_{i}) \\
        & \ \ \ + P(x_{m}) \sum_{i=1}^{n} \rho_{m,i}(f_{m}-f_{i}) )
    \end{aligned}
\end{multline}
where $\rho_m$ is the estimation of $f_m$, i.e., $\rho_m = {\textstyle \sum_{i=1}^{n}}e^{-\frac{1}{2 h^{2}}\left(f_m-f_i\right)^{2}}$.

\subsubsection{Optimization under Inequality Constraints}
The constrained optimization is handled by the method of Lagrange multipliers with the Karush-Kuhn-Tucker (KKT) conditions.

The Lagrangian $L(F,\ \lambda)$ for our problem is given by:
\begin{equation}
  L(F,\ \lambda) = loss + \lambda g(F)
\end{equation}
where $\lambda$ is the Lagrange multiplier associated with the inequality constraint.

The associated KKT conditions are as follows:
\begin{equation}
  \begin{array}{l}
    \nabla_{F} L(F,\ \lambda)=\frac{\partial loss}{\partial F}+\lambda \nabla g(F)=0 \\
    \text { s.t. } g(F) \leq 0,\ \lambda \geqslant 0,\ \lambda g(F)=0
  \end{array}    
\end{equation}
With the slope of gradient clarified in the previous section, we can perform gradient descent using a properly chosen optimization algorithm, such as Sequential Least Squares Quadratic Programming (SLSQP).

\subsection{Normalization Equivalence}
Theoretically speaking, our approach will not guarantee the obtained $F$ falling into a specific range. Practical reasoning concludes that $F$ should be appropriately scaled in order not to jeopardize the impact of $\kappa$ in Formula \ref{eq:k}.

In this section, we prove that the scaling of $F$ is not necessary in DaNNG as long as it maintains symmetric about $\kappa$.

\begin{proof}
A formal restatement of this matter can be summarized as follows:
\begin{equation} \label{eq:normalization1}
    \begin{aligned}
        \forall \ F_1=g_1(F&),\ F_2=g_2(F),\ \kappa \in \mathbb{R} \\
        \exists \ \eta_1, \eta_2,   \text{ s.t. } & (1-\eta_1)\kappa + \eta_1 F_1\\
        \equiv &(1-\eta_2)\kappa + \eta_1 F_2
    \end{aligned}
\end{equation}
where $g(\cdot)$ denotes the normalization of mapping $F$ into range $(Min,\ Max)$, $g(F)= \frac{{(F - f_{min}) \times (Max - Min)}}{{f_{max} - f_{min}}} + Min$, s.t. $(Max_i \neq Min_i) \wedge (\frac{Max_i + Min_i}{2} = \kappa)$.

\begin{theorem} \label{bitheorem}
    Bijective Linear Transformation -- the normalization of $g(\cdot): \mathbb{R}^{n\times 1} \rightarrow \mathbb{R}^ {n\times 1}$ indicates a self-explanatory bijective linear transformation between vector spaces that are both injective (one-to-one) and surjective (onto), implying that it has a unique inverse $\text{iff. } Max \neq Min$.
\end{theorem}

Because of Theorem \ref{bitheorem}, the identity in formula \ref{eq:normalization1} becomes:
\begin{equation} \label{eq:normalization2}
    \eta_{1}g_1(g_2^{-1}(F_2))-\eta_{2}F_2+\left(\eta_{2}-\eta_{1}\right)\kappa\equiv 0
\end{equation}

With $g(F)= \frac{{(F - f_{min}) \times (Max - Min)}}{{f_{max} - f_{min}}} + Min$ and $F_2 \in \mathbb{R}^{n\times 1}$, we have:
\begin{equation} \label{eq:normalization3}
    \left\{
    \begin{aligned}
    \frac{{Max_1 - Min_1}}{{Max_2 - Min_2}} \times \eta_1 - \eta_2 &= 0 \\
    (Min_1 - \kappa)\eta_1 + (\kappa - Min_2)\eta_2 &= 0
    \end{aligned}
    \right.
\end{equation}
where $Max_i$ and $Min_i$ represent the target range of the $i_{th}$ normalization.

To determine whether the above homogeneous system have solutions for $\eta_1$ and $\eta_2$, we rewrite the corresponding coefficient matrix:
\begin{equation} \label{eq:normalization5}
    \begin{bmatrix}
        1 & -\frac{1}{\alpha}\\
        0 & \alpha (\kappa - Min_2) - (\kappa - Min_1)
    \end{bmatrix}   
\end{equation}

in which $\alpha$ represents the fraction of ranges (i.e., $\alpha = \frac{Max_1-Min_1}{Max_2-Min_2}$). The coefficient matrix has rank 1 $\text{iff. } (Max_i \neq Min_i) \wedge (\frac{Max_i + Min_1}{2} = \kappa)$.
\end{proof}

Thus, there exist solutions as $\eta_{2}=\alpha \times \eta_{1}$, s.t. $(\alpha = \frac{Max_1-Min_1}{Max_2-Min_2}) \wedge (Max_i \neq Min_i)$ for any pair of $\kappa$-symmetric normalizations $g_1(\cdot)$ and $g_2(\cdot)$ to be equivalent. In other words, although we scale $F$ to $(\frac{1}{2}\kappa, \frac{3}{2}\kappa)$ to ensure consistent impacts across different $\kappa$, the flexibility of the model is enriched by automatically incorporating various ranges through the trade-off parameter $\eta$ without requiring any additional procedures.

\section{Evaluations} \label{sec:evaluations}

\subsection{Experiment Settings}
\begin{table}[ht]
    \begin{center}
    \begin{tabular}{crrr}
    \hline
    \textbf{Dataset} & \textbf{Instances} & \textbf{Classes} & \textbf{Dimensions} \\
    \hline
    WDBC & 569 & 2 & 30 \\
    BCC & 116 & 2 & 9 \\
    WPBC & 198 & 2 & 30 \\
    Heart Disease & 303 & 4 & 13 \\
    Abalone & 4177 & 28 & 8 \\
    Glass & 214 & 6 & 9 \\
    Zoo & 101 & 7 & 16 \\
    Glioma & 839 & 2 & 23 \\
    Diabetes & 768 & 2 & 7 \\
    Dermatology & 366 & 6 & 34 \\
    Wine & 178 & 3 & 13 \\
    German Credit & 1000 & 2 & 20 \\
    \hline
    \end{tabular}
    \end{center}
    \caption{Statistics of included datasets.}
    \label{tab:datasets}
\end{table}

\begin{table*}[t]
  \small 
  \begin{tabular}{crrrrrrrrrrr}
  \hline
    & PL-kNN & SMKNN & LMKNN & LV-kNN & AKNNG & MAKNNG & Centered & Mutual & kNNG & DaNNG & DaNNG\\
    \textbf{Dataset} & & & & & & & kNNG & kNNG & & (plain) & \\
  \hline
  WDBC & 0.9086 & 0.8999 & 0.8981 & 0.9174 & 0.9192 & 0.9210 & 0.9050 & 0.9192 & 0.9157 & \textbf{0.9315} & 0.9294 \\
  BCC & 0.3652 & 0.2705 & 0.3515 & 0.3576 & 0.4485 & 0.4568 & 0.2720 & 0.4008 & 0.3750 & 0.3801 & \textbf{0.5515} \\
  WPBC & 0.6989 & 0.7092 & \textbf{0.7639} & 0.7489 & 0.7334 & 0.7334 & 0.6987 & 0.7237 & 0.7337 & 0.7578 & 0.7589 \\
  Heart Disease & 0.3242 & 0.2951 & 0.3302 & 0.3291 & 0.2555 & 0.2626 & 0.3516 & 0.3423 & 0.3143 & 0.3434 & \textbf{0.4016} \\
  Abalone & 0.2217 & 0.2274 & 0.2200 & 0.2459 & 0.2049 & 0.1994 & 0.2298 & 0.2298 & 0.2286 & \textbf{0.2906} & 0.2726 \\
  Glass & 0.5167 & 0.3758 & 0.2641 & 0.4251 & 0.5216 & 0.5262 & 0.4805 & 0.5301 & 0.4290 & 0.4290 & \textbf{0.5688} \\
  Zoo & 0.9400 & 0.9100 & 0.7309 & 0.8500 & 0.9400 & \textbf{0.9600} & 0.7800 & 0.8200 & 0.7800 & 0.9400 & 0.9567 \\
  Glioma & 0.8296 & 0.8022 & 0.8081 & \textbf{0.8535} & 0.8152 & 0.8032 & 0.8163 & 0.8188 & 0.8475 & 0.8475 & 0.8487 \\
  Diabetes & 0.6420 & 0.6536 & 0.6510 & 0.6991 & 0.6692 & 0.6588 & 0.6561 & 0.6925 & \textbf{0.7096} & \textbf{0.7096} & \textbf{0.7096} \\
  Dermatology & 0.8013 & \textbf{0.8637} & 0.7685 & 0.7736 & 0.7956 & 0.7983 & 0.7957 & 0.7872 & 0.7681 & 0.7953 & 0.8011 \\
  Wine & 0.7261 & 0.6690 & 0.5490 & 0.7026 & 0.7363 & 0.7418 & 0.6278 & 0.6791 & 0.6912 & 0.7186 & \textbf{0.7869} \\
  German Credit & 0.7030 & 0.6940 & 0.7000 & 0.6990 & 0.6940 & 0.6790 & 0.6400 & 0.7020 & 0.7080 & \textbf{0.7200} & 0.713 \\
  \hline
  \textbf{Average} & 0.6398 & 0.6142 & 0.5863 & 0.6335 & 0.6445 & 0.6450 & 0.6045 & 0.6371 & 0.6251 & 0.6552 & \textbf{0.6916} \\
  \hline
  \end{tabular}
  \caption{Accuracy of 9 SOTAs and DaNNG on 12 Datasets.}
  \label{tab:performance}
\end{table*}

\begin{table*}[t]
  \small 
  \begin{tabular}{crrrrrrrrrrr}
    \hline
    & PL-kNN & SMKNN & LMKNN & LV-kNN & AKNNG & MAKNNG & Centered & Mutual & kNNG & DaNNG & DaNNG\\
    \textbf{Dataset} & & & & & & & kNNG & kNNG & & (plain) & \\
  \hline
  WDBC & 0.8966 & 0.9001 & 0.9025 & 0.9036 & 0.8858 & 0.8859 & 0.8752 & 0.8722 & 0.8987 & \textbf{0.9087} & 0.8991 \\
  BCC & 0.2637 & 0.2840 & 0.3285 & 0.3367 & 0.2014 & 0.2551 & 0.1848 & 0.2212 & 0.2628 & 0.2628 & \textbf{0.3713} \\
  WPBC & 0.4539 & 0.4167 & 0.5343 & 0.5519 & 0.4555 & 0.4694 & 0.5480 & 0.4305 & 0.4534 & 0.4627 & \textbf{0.5792} \\
  Heart Disease & 0.1927 & 0.2090 & 0.1690 & 0.1951 & \textbf{0.2798} & 0.2014 & 0.1844 & 0.1732 & 0.1785 & 0.2176 & 0.2622 \\
  Abalone & 0.1493 & 0.1606 & 0.1320 & 0.1320 & 0.1240 & 0.0882 & 0.0544 & 0.0544 & 0.1526 & 0.1691 &\textbf{0.1719} \\
  Glass & 0.2333 & 0.2931 & 0.2843 & 0.2790 & 0.2800 & 0.2767 & 0.1948 & 0.2033 & 0.2306 & 0.2482 & \textbf{0.3355} \\
  Zoo & 0.6246 & 0.6714 & 0.8607 & \textbf{0.9190} & 0.6262 & 0.9057 & 0.8395 & 0.6468 & 0.7432 & 0.8588 & 0.9089 \\
  Glioma & 0.4970 & 0.4833 & 0.4865 & 0.4773 & 0.4877 & 0.4914 & 0.4811 & 0.4829 & 0.4987 & 0.4987 &\textbf{0.5071} \\
  Diabetes & \textbf{0.6544} & 0.6366 & 0.6163 & 0.6095 & 0.5971 & 0.5784 & 0.4240 & 0.4084 & 0.6381& 0.6381 & 0.6381 \\
  Dermatology & 0.7356 & 0.7618 & 0.7656 & 0.7703 & 0.7655 & 0.7652 & \textbf{0.8381} & 0.6562 & 0.6954 & 0.7424 & 0.7868 \\
  Wine & 0.4056 & 0.3601 & 0.4170 & 0.4055 & \textbf{0.4658} & 0.4550 & 0.3807 & 0.3128 & 0.3792 & 0.3685 & 0.4414 \\
  German Credit & 0.5781 & 0.5770 & 0.5839 & 0.5719 & 0.5950 & 0.4954 & 0.4337 & 0.4113 & 0.5633 & \textbf{0.6050}& 0.5950 \\
  \hline
  \textbf{Average} & 0.4737 & 0.4795 & 0.5067 & 0.5126 & 0.4803 & 0.4890 & 0.4532 & 0.4061 & 0.4745 & 0.4984 & \textbf{0.5414} \\
  \hline
  \end{tabular}
  \caption{F1-Score of 9 SOTAs and DaNNG on 12 Datasets.}
  \label{tab:f1}
\end{table*}

We evaluate DaNNG's performance on both binary classification and multi-class classification tasks against nine state-of-the-art algorithms, including four adaptive kNN models, i.e., PL-kNN \cite{plknn}, SMKNN \cite{smknn}, LMKNN \cite{smknn}, LV-kNN \cite{lvknn} and five graph-based kNN models i.e., the Centered kNNG \cite{centered}, AKNNG \cite{aknng}, MAKNNG \cite{aknng}, mutual kNNG \cite{mutual_knng}, and plain kNNG\footnote{Model abbreviation is consistent with how it's described in original works, and all implementations of both SOTAs and DaNNG are publicly available.}. Additionally, to test the effectiveness of the proposed adaption criteria, we also include a version of DaNNG where we set aside the constraints during optimization, denoted as DaNNG(plain). Table \ref{tab:datasets} provides information on twelve included datasets obtained from UCI Machine Learning Lab, except for the Diabetes dataset \cite{smithUsing1988} which was sourced from Kaggle. 

To ensure objectivity prior to assessment, rigorous preprocessing has been applied to the data, and a uniform pre-configuration has been implemented for all algorithms. Detailed data sources and preprocessing methodologies are presented in \textit{Appendices}.

\subsection{Overall Performance}

The performance matrix in Table \ref{tab:performance} showcases the accuracy of different algorithms on each dataset. Across multiple datasets, DaNNG consistently demonstrates strong performance. Notably, on five datasets: BCC, Heart Disease, Glass, Diabetes, and Wine, DaNNG outperforms all other competing algorithms. Our model also exhibits the highest average accuracy among all twelve datasets. We also provide the marco F1-score to assess overall effectiveness of DaNNG in scenarios with multi-class and class imbalanced classifications. Table \ref{tab:f1} shows that DaNNG also consistently outperforms competing SOTAs with resepect to F1-score.

Significantly, while DaNNG(plain) demonstrates substantial robustness across multiple datasets, augmenting the optimization process with the proposed constraints further enhances our model's performance. The improvements achieved by DaNNG, in terms of both accuracy and F1-score in comparison to DaNNG(plain), serve as validation for the effectiveness of the criteria introduced in Section \ref{sec:criteria}.

\section{Discussion}

In this section, we conduct a detailed examination of our model's performance, with particular attention to its comparison with kNNG. Our model introduces distribution information into the graph construction of kNNG and, significantly, outperforms kNNG in 11 out of 12 datasets. This strong performance advantage sets the stage for an in-depth comparison that underscores the merits of our approach in enhancing classification outcomes.

\subsection{Sensitivity of the Trade-Off Parameter}

\begin{figure}[ht]
  \centering
  \includegraphics[width=0.9\columnwidth]{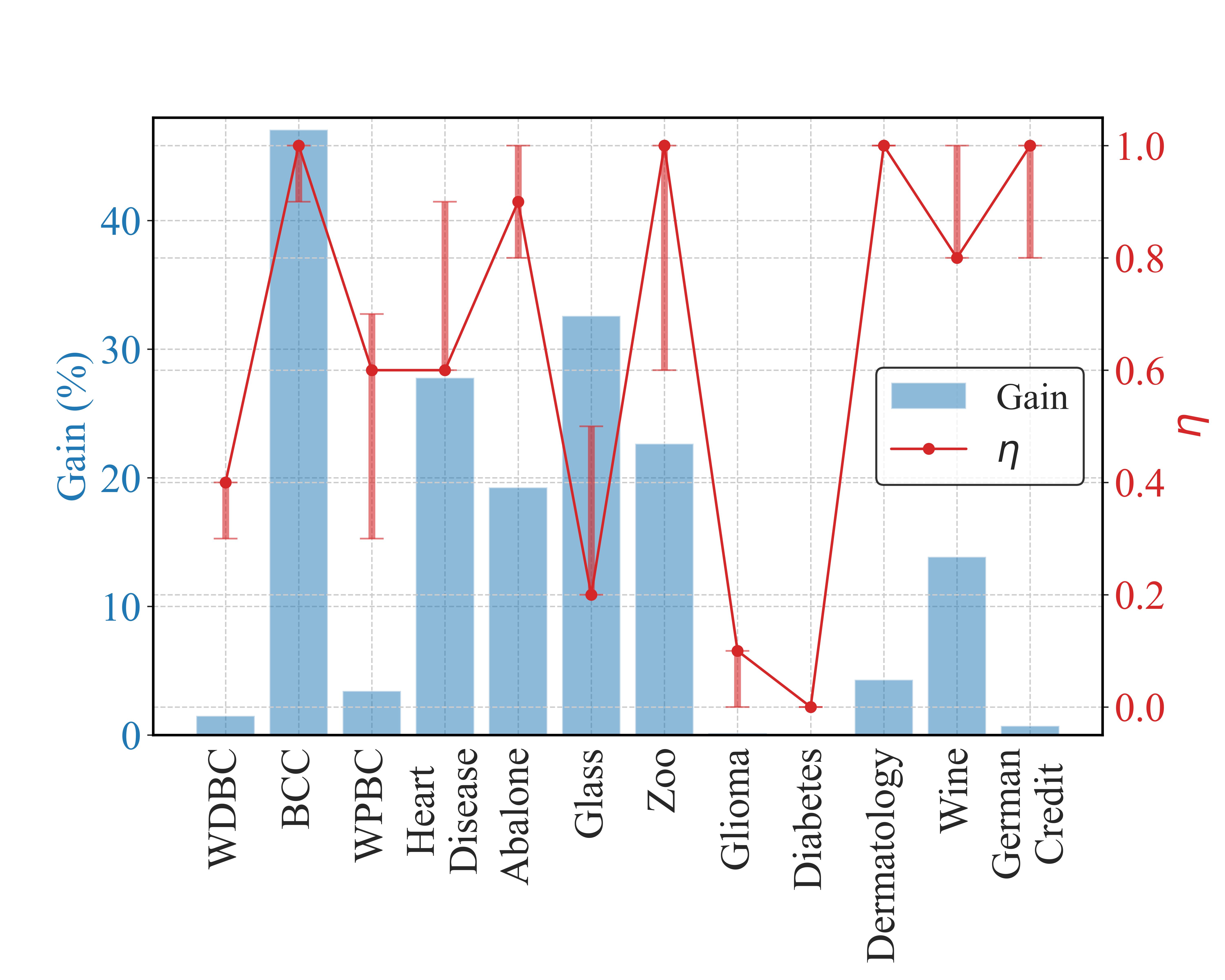} 
  \caption{Performance gain of DaNNG concerning kNNG and the corresponding optimal trade-off $\eta$ with error bar.}
  \Description{Consistent gains of performance observed in DaNNG with respect to kNNG for 10 Datasets}
  \label{fig:gain_of_eta}
\end{figure}

The trade-off $\eta$ is a crucial hyper-parameter of DaNNG that balances the influence of the fitness kernel and traditional \(k\). A higher $\eta$ emphasizes the importance of the fitness kernel and hence the effectiveness of our proposal, while a lower $\eta$ indicates possible unnecessary.

Figure \ref{fig:gain_of_eta} shows the percentage of accuracy gain of DaNNG regarding kNNG \(\left(gain = \frac{{accuracy_\text{DaNNG} - accuracy_\text{kNNG}}}{{accuracy_\text{kNNG}}}\right)\) under the optimal values of $\eta$ in Section \ref{sec:evaluations}. From Figure \ref{fig:gain_of_eta}, we observe a notable trend that as long as $\eta$ is not close to zero, the accuracy gains consistently, with a maximum of 47.07\% for BCC.

This finding underscores the significance of incorporating distribution approximation through the fitness kernel in our model. The higher $\eta$ values indicate a stronger reliance on distribution-informed adaption, leading to improved classification performance.

\subsection{Significant Improvement on Borderline Samples}

As has been illustrated in Section \ref{sec:matters}, the mechanism behind the performance improvement of distribution-informed adaptive-k can be credited to the behavior of borderline samples. To evaluate the impact of our model on borderline samples, we analyze the performance gain of DaNNG specifically on these samples. 
\begin{figure}[t]
    \centering
    \includegraphics[width=0.9\columnwidth]{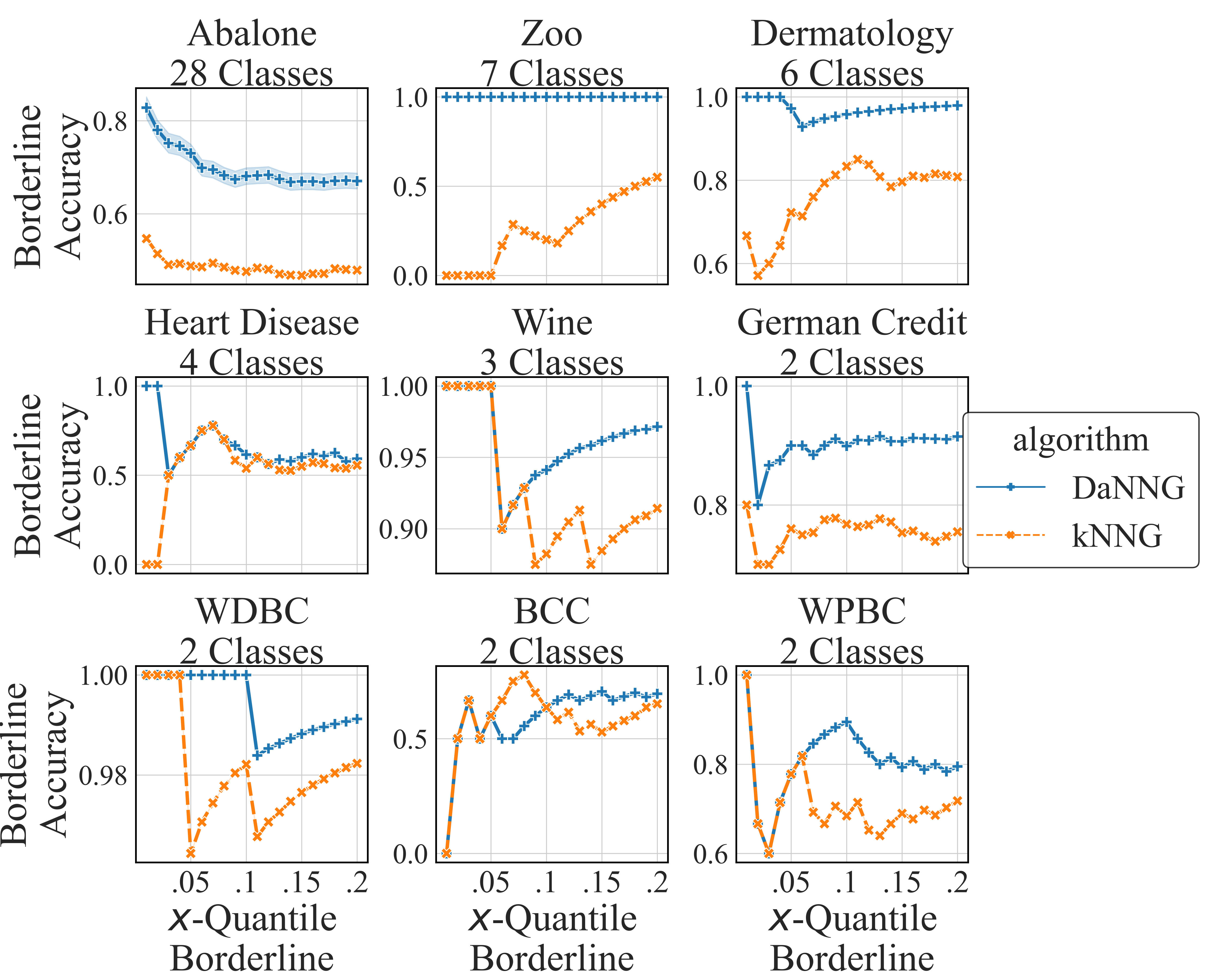}
    \caption{Borderline-sample accuracies of DaNNG and kNNG.}
    \Description{Accuracies of DaNNG on borderline samples are significantly higher than kNNG.}
    \label{fig:borderline_accuracies}
\end{figure}

We have computed the subsets of 1-Quantile Borderline to 20-Quantile Borderline for the aforementioned 9 datasets where $\eta > 0.2$. The accuracies on those borderline-sample subsets are shown in Figure \ref{fig:borderline_accuracies}. We observe significant improvements in borderline accuracies, especially in datasets involving a large number of classes. Improvements on binary classification are also consistently stable and notable, particularly with relatively larger datasets (i.e., WDBC and German Credit).

The considerable enhancement of borderline-sample performance of DaNNG provides insight for both the advancement of multi-class classifications and the dilemma of ambiguous decision boundaries.

\section{Conclusion}
In this article, we introduce DaNNG, a graph-based adaptive kNN model that integrates distribution approximation into the determination of $k$ values. Through rigorous analysis, we establish that our approach generates a heterogeneous graph, significantly enhancing borderline accuracies and thereby improving overall model performance and robustness. Empirical evaluations demonstrate that DaNNG outperforms state-of-the-art adaptive kNN and graph-based kNN algorithms in various real-world scenarios. Notably, DaNNG not only achieves satisfied performance but also partially bypasses the necessity for normalization via the trade-off parameter $\eta$, enriching model flexibility. The model's adaptability and context-awareness position it as a potential solution for a wide range of real-world applications, particularly in class-imbalanced and multi-class scenarios.

\section{Appendices}

\bibliographystyle{ACM-Reference-Format}
\bibliography{sample-base}

\appendix

\section{Data Appendix}
The included datasets encompass a variety of domains and data complexities, ranging from medical diagnoses (e.g., Heart Disease, Diabetes, and Dermatology) to physical properties (e.g., Abalone, Glass, and Wine). The number of instances varies from small datasets like Zoo to larger datasets like Abalone. The model's performance is evaluated on these datasets to assess its adaptability and generalization capability across different real-world scenarios. Detailed data resources are in Table \ref{tab:sup_datasets}.

\begin{table}[ht]
    \begin{center}
    \begin{tabular}{cc}
    \hline
    \textbf{Dataset} & \textbf{Resource} \\
    \hline
    WDBC \cite{wdbc} & UCI \\
    BCC \cite{bcc} & UCI \\
    WPBC \cite{wpbc} & UCI \\
    Heart Disease \cite{heart_disease} & UCI \\
    Abalone \cite{abalone} & UCI \\
    Glass \cite{glass} & UCI \\
    Zoo \cite{zoo} & UCI \\
    Glioma \cite{glioma} & UCI \\
    Diabetes \cite{diabetes} & Kaggle \\
    Dermatology \cite{dermatology} & UCI \\
    Wine \cite{wine} & UCI \\
    German Credit \cite{german_credit} & UCI \\
    \hline
    \end{tabular}
    \end{center}
    \caption{Resources of included datasets.}
    \label{tab:sup_datasets}
\end{table}

\section{Code Appendix}
Our evaluaions are conducted under Python3 with package requirement specified in Table \ref{tab:requirements}. All 10 algorithms (including DaNNG) are implemented as a sklearn classifier with functions, $fit(\cdot)$ and $predict(\cdot)$ \footnote{All implementations are available in: https://github.com/4alexmin/knnsotas}.

\begin{table}[h]
    \begin{center}
    \begin{tabular}{cc}
    \hline
    \textbf{Package} & \textbf{Version}\\
    \hline
    networkx & 2.8.8  \\
    numpy & 1.21.2 \\
    pandas & 1.3.3 \\
    scikit-learn & 1.1.3 \\
    scipy & 1.7.1 \\
    prda & 1.1.0 \\
    \hline
    \end{tabular}
    \end{center}
    \caption{Python package requirements of evaluaions.}
    \label{tab:requirements}
\end{table}

Before conducting the experiments, datasets are preprocessed to ensure uniformity. The following preprocessing steps are applied:

\begin{itemize}
    \item Samples without the prediction target are dropped from datasets.
    \item Samples with missing values in features are filled with the mean value of the corresponding feature.
    \item Numerical features are normalized before fitting, and categorical features are preprocessed with one-hot encoder.
\end{itemize}

Models are pre-configured as follows:

\begin{itemize}
    \item Models which require a pre-defined $k$ are set to 10, and those that require a pre-defined $k$ boundary ($k_{min}$ and $k_{max}$) are set to 5 and 15, respectively.
    \item Graph-based algorithms, including our DaNNG, use the majority vote strategy to make predictions, meaning that the majority class of neighbors of the closest node for each test sample is applied as the prediction.
    \item In DaNNG, the trade-off parameter $\eta$ is set to range $(0, 1)$ with an interval of 0.1 in order to assess the degree of importance of $F$, and the KDE bandwidth $h$ is set to 0.5.
    \item The performance is evaluated by measuring prediction accuracy under 10-fold cross-validation.
\end{itemize}

\end{document}